\documentclass[sigconf]{acmart}

\copyrightyear{2022}
\acmYear{2022}
\setcopyright{rightsretained}
\acmConference[KDD '22]{Proceedings of the 28th ACM SIGKDD Conference on Knowledge Discovery and Data Mining}{August 14--18, 2022}{Washington, DC, USA}
\acmBooktitle{Proceedings of the 28th ACM SIGKDD Conference on Knowledge Discovery and Data Mining (KDD '22), August 14--18, 2022, Washington, DC, USA}
\acmDOI{10.1145/3534678.3539043}
\acmISBN{978-1-4503-9385-0/22/08}

\usepackage{hyperref}
\usepackage{multirow}

\usepackage{textcomp}
\newcommand{\textapproxA}{\raisebox{0.5ex}{\texttildelow}}

\begin{document}
%

\title{DocLayNet: A Large Human-Annotated Dataset for Document-Layout Analysis}

\author{Birgit Pfitzmann}
\affiliation{%
  \institution{IBM Research}
  \city{Rueschlikon}
  \country{Switzerland}}
\email{bpf@zurich.ibm.com}

\author{Christoph Auer}
\affiliation{%
  \institution{IBM Research}
  \city{Rueschlikon}
  \country{Switzerland}}
\email{cau@zurich.ibm.com}

\author{Michele Dolfi}
\affiliation{%
  \institution{IBM Research}
  \city{Rueschlikon}
  \country{Switzerland}}
\email{dol@zurich.ibm.com}

\author{Ahmed S. Nassar}
\affiliation{%
  \institution{IBM Research}
  \city{Rueschlikon}
  \country{Switzerland}}
\email{ahn@zurich.ibm.com}

\author{Peter Staar}
\affiliation{%
  \institution{IBM Research}
  \city{Rueschlikon}
  \country{Switzerland}}
\email{taa@zurich.ibm.com}



\begin{abstract}
Accurate document layout analysis is a key requirement for high-quality PDF document conversion. With the recent availability of public, large ground-truth datasets such as PubLayNet and DocBank, deep-learning models have proven to be very effective at layout detection and segmentation. While these datasets are of adequate size to train such models, they severely lack in layout variability since they are sourced from scientific article repositories such as PubMed and arXiv only.  Consequently, the accuracy of the layout segmentation drops significantly when these models are applied on more challenging and diverse layouts. In this paper, we present \textit{DocLayNet}, a new, publicly available, document-layout annotation dataset in COCO format.  It contains 80863 manually annotated pages from diverse data sources to represent a wide variability in layouts. For each PDF page, the layout annotations provide labelled bounding-boxes with a choice of 11 distinct classes.  DocLayNet also provides a subset of double- and triple-annotated pages to determine the inter-annotator agreement.  In multiple experiments, we provide baseline accuracy scores (in mAP) for a set of popular object detection models. We also demonstrate that these models fall approximately 10\% behind the inter-annotator agreement. Furthermore, we provide evidence that DocLayNet is of sufficient size. Lastly, we compare models trained on PubLayNet, DocBank and DocLayNet, showing that layout predictions of the DocLayNet-trained models are more robust and thus the preferred choice for general-purpose document-layout analysis.
\end{abstract}

\ccsdesc[500]{Information systems~Document structure}
\ccsdesc[500]{Applied computing~Document analysis}
\ccsdesc[500]{Computing methodologies~Machine learning}
\ccsdesc[500]{Computing methodologies~Computer vision}
\ccsdesc[300]{Computing methodologies~Object detection}

\keywords{PDF document conversion, layout segmentation, object-detection, data set, Machine Learning}

\maketitle

\begin{figure}[t!]
  \center
  \includegraphics[width=\linewidth]{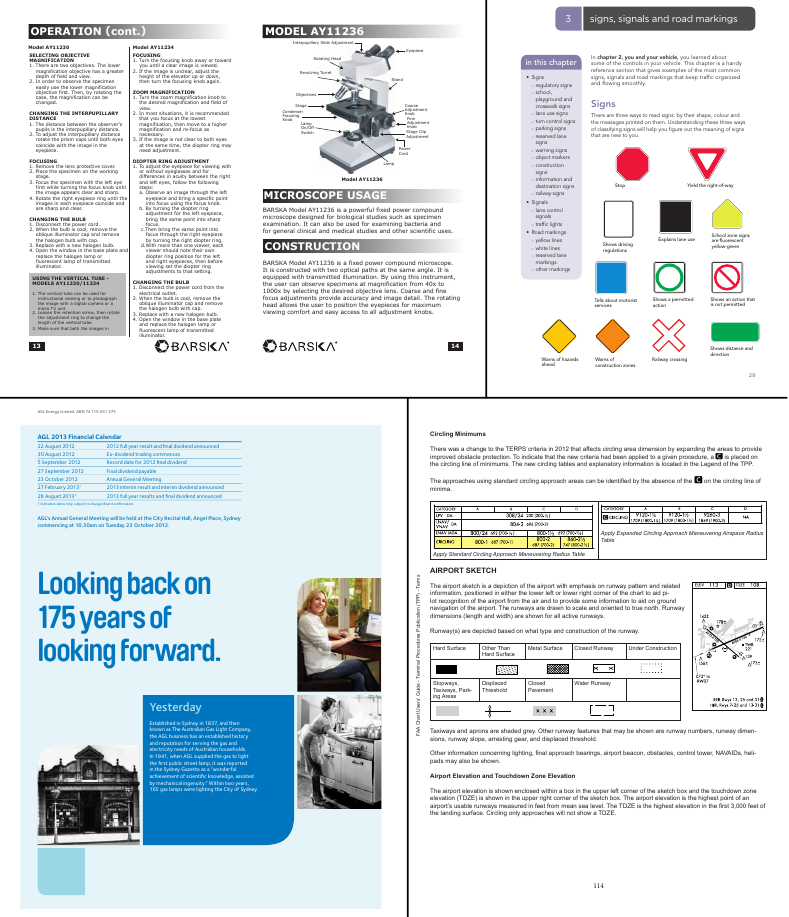}
\caption{\label{fig:complexpages} Four examples of complex page layouts across different document categories}
\end{figure}

\section{Introduction}\label{sec:Intro}
\vspace*{3.5mm}

Despite the substantial improvements achieved with machine-learning (ML) approaches and deep neural networks in recent years, document conversion remains a challenging problem, as demonstrated by the numerous public competitions held on this topic~\cite{ICDAR2013,ICDAR2017,ICDAR2019,Jimeno-etal-2021-PubLayNetCompetition}. The challenge originates from the huge variability in PDF documents regarding layout, language and formats (scanned, programmatic or a combination of both).  Engineering a single ML model that can be applied on all types of documents and provides high-quality layout segmentation remains to this day extremely challenging~\cite{SegNotDead}. To highlight the variability in document layouts, we show a few example documents from the DocLayNet dataset in Figure~\ref{fig:complexpages}. 

A key problem in the process of document conversion is to understand the structure of a single document page, i.e. which segments of text should be grouped together in a unit. To train models for this task, there are currently two large datasets available to the community, PubLayNet~\cite{Zhong-etal-2019-PubLayNet} and DocBank~\cite{Li-etal-2020-docbank}. They were introduced in 2019 and 2020 respectively and significantly accelerated the implementation of layout detection and segmentation models due to their sizes of 300K and 500K ground-truth pages. These sizes were achieved by leveraging an automation approach.
The benefit of automated ground-truth generation is obvious: one can generate large ground-truth datasets at virtually no cost. However, the automation introduces a constraint on the variability in the dataset, because corresponding structured source data must be available. PubLayNet and DocBank were both generated from scientific document repositories (PubMed and arXiv), which provide XML or \LaTeX\,\, sources. Those scientific documents present a limited variability in their layouts, because they are typeset in uniform templates provided by the publishers. Obviously, documents such as technical manuals, annual company reports, legal text, government tenders, etc. have very different and partially unique layouts. As a consequence, the layout predictions obtained from models trained on PubLayNet or DocBank is very reasonable when applied on scientific documents. However, for more \textit{artistic} or \textit{free-style} layouts, we see sub-par prediction quality from these models, which we demonstrate in Section~\ref{sec:results}.

In this paper, we present the DocLayNet dataset. It provides page-by-page layout annotation ground-truth using bounding-boxes for 11 distinct class labels on 80863 unique document pages, of which a fraction carry double- or triple-annotations. DocLayNet is similar in spirit to PubLayNet and DocBank and will likewise be made available to the public\footnote{\url{https://developer.ibm.com/exchanges/data/all/doclaynet}} in order to stimulate the document-layout analysis community. It distinguishes itself in the following aspects:

\begin{enumerate}
\item \textit{Human Annotation}: In contrast to PubLayNet and DocBank, we relied on human annotation instead of automation approaches to generate the data set.
\item \textit{Large Layout Variability}: We include diverse and complex layouts from a large variety of public sources.
\item \textit{Detailed Label Set}: We define 11 class labels to distinguish layout features in high detail. PubLayNet provides 5 labels; DocBank provides 13, although not a superset of ours.
\item \textit{Redundant Annotations}: A fraction of the pages in the DocLayNet data set carry more than one human annotation. This enables experimentation with annotation uncertainty and quality control analysis.
\item \textit{Pre-defined Train-, Test- \& Validation-set}: Like DocBank, we provide fixed train-, test- \& validation-sets to ensure proportional representation of the 
class-labels. Further, we prevent leakage of unique layouts across sets, which has a large effect on model accuracy scores.    
\end{enumerate}

All aspects outlined above are detailed in Section~\ref{sec:DLN}. In Section~\ref{sec:DAC}, we will elaborate on how we designed and executed this large-scale human annotation campaign. We will also share key insights and lessons learned that might prove helpful for other parties planning to set up annotation campaigns.

In Section~\ref{sec:results}, we will present baseline accuracy numbers for a variety of object detection methods (Faster R-CNN, Mask R-CNN and YOLOv5) trained on DocLayNet. We further show how the model performance is impacted by varying the DocLayNet dataset size, reducing the label set and modifying the train/test-split. Last but not least, we compare the performance of models trained on PubLayNet, DocBank and DocLayNet and demonstrate that a model trained on DocLayNet provides overall more robust layout recovery.

\section{Related Work}\label{sec:RW}
\vspace*{3.5mm}

While early approaches in document-layout analysis used rule-based algorithms and heuristics~\cite{Ahmad2016InformationEF}, the problem is lately addressed with deep learning methods. The most common approach is to leverage object detection models~\cite{RCNN, Fast-RCNN, FasterRCNN, MaskRCNN, Yolov5, DETR, EfficientDet}. In the last decade, the accuracy and speed of these models has increased dramatically. Furthermore, most state-of-the-art object detection methods can be trained and applied with very little work, thanks to a standardisation effort of the ground-truth data format~\cite{Lin-etal-2014-COCO} and common deep-learning frameworks~\cite{detectron2}. Reference data sets such as PubLayNet~\cite{Zhong-etal-2019-PubLayNet} and DocBank provide their data in the commonly accepted COCO format~\cite{Lin-etal-2014-COCO}.

Lately, new types of ML models for document-layout analysis have emerged in the community~\cite{Livathinos-et-al-2021-Seq2Seq, Xu-etal-2020-LayoutLM,Li-etal-2021-VTLayout, Zhang-etal-2021-VSR}. These models do not approach the problem of layout analysis purely based on an image representation of the page, as computer vision methods do. Instead, they combine the text tokens and image representation of a page in order to obtain a segmentation. While the reported accuracies appear to be promising, a broadly accepted data format which links geometric and textual features has yet to establish.

\section{The DocLayNet Dataset}\label{sec:DLN}
\vspace*{3.5mm}

DocLayNet contains 80863 PDF pages. Among these, 7059 carry two instances of human annotations, and 1591 carry three. This amounts to 91104 total annotation instances. 
The annotations provide layout information in the shape of labeled, rectangular bounding-boxes. We define 11 distinct labels for layout features, namely \textit{Caption}, \textit{Footnote}, \textit{Formula}, \textit{List-item}, \textit{Page-footer}, \textit{Page-header}, \textit{Picture}, \textit{Section-header}, \textit{Table}, \textit{Text}, and \textit{Title}. Our reasoning for picking this particular label set is detailed in Section~\ref{sec:DAC}.

In addition to open intellectual property constraints for the source documents, we required that the documents in DocLayNet adhere to a few conditions. Firstly, we kept scanned documents to a minimum, since they introduce difficulties in annotation (see Section~\ref{sec:DAC}). As a second condition, we focussed on medium to large documents ($>10$ pages) with technical content, dense in complex tables, figures, plots and captions. Such documents carry a lot of information value, but are often hard to analyse with high accuracy due to their challenging layouts. Counterexamples of documents not included in the dataset are receipts, invoices, hand-written documents or photographs showing ``text in the wild".


\begin{table*}[t!]
\centering
\caption{\label{tab:DLN}DocLayNet dataset overview. Along with the frequency of each class label, we present the relative occurrence (as \% of row ``Total'') in the train, test and validation sets. The inter-annotator agreement is computed as the mAP@0.5-0.95 metric between pairwise annotations from the triple-annotated pages, from which we obtain accuracy ranges.
}
\vspace*{3.5mm}
\begin{tabular}{|l|r|rrr|rrrrrrr|}
\hline
          &         &\multicolumn{3}{|c|}{\% of Total}            &  \multicolumn{7}{c|}{triple inter-annotator \:mAP @ 0.5-0.95 (\%)}   \\
class label          & Count  & Train  & Test   & Val       & All      & Fin   & Man   & Sci   & Law   & Pat   & Ten   \\ \hline
Caption         &  22524 &  2.04  &  1.77  &  2.32     & 84-89    & 40-61 & 86-92 & 94-99 & 95-99 & 69-78 & n/a   \\
Footnote        &   6318 &  0.60  &  0.31  &  0.58     & 83-91    & n/a   & 100   & 62-88 & 85-94 & n/a   & 82-97 \\
Formula         &  25027 &  2.25  &  1.90  &  2.96     & 83-85    & n/a   & n/a   & 84-87 & 86-96 & n/a   & n/a   \\
List-item       & 185660 & 17.19  & 13.34  & 15.82     & 87-88    & 74-83 & 90-92 & 97-97 & 81-85 & 75-88 & 93-95 \\
Page-footer     &  70878 &  6.51  &  5.58  &  6.00     & 93-94    & 88-90 & 95-96 & 100   & 92-97 & 100   & 96-98 \\
Page-header     &  58022 &  5.10  &  6.70  &  5.06     & 85-89    & 66-76 & 90-94 & 98-100& 91-92 & 97-99 & 81-86 \\
Picture         &  45976 &  4.21  &  2.78  &  5.31     & 69-71    & 56-59 & 82-86 & 69-82 & 80-95 & 66-71 & 59-76 \\
Section-header  & 142884 & 12.60  & 15.77  & 12.85     & 83-84    & 76-81 & 90-92 & 94-95 & 87-94 & 69-73 & 78-86 \\
Table           &  34733 &  3.20  &  2.27  &  3.60     & 77-81    & 75-80 & 83-86 & 98-99 & 58-80 & 79-84 & 70-85 \\
Text            & 510377 & 45.82  & 49.28  & 45.00     & 84-86    & 81-86 & 88-93 & 89-93 & 87-92 & 71-79 & 87-95 \\
Title           &   5071 &  0.47  &  0.30  &  0.50     & 60-72    & 24-63 & 50-63 & 94-100& 82-96 & 68-79 & 24-56 \\ \hline
Total           &1107470 & 941123 &  99816 &  66531    & 82-83    & 71-74 & 79-81 & 89-94 & 86-91 & 71-76 & 68-85 \\ \hline
\end{tabular}
\end{table*}



\begin{figure}[t!]
 \center
 \includegraphics[width=\linewidth]{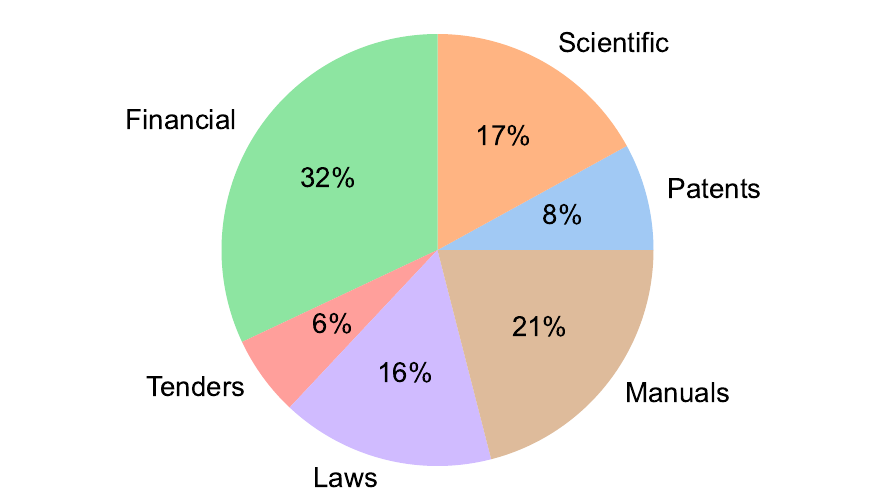}
\caption{Distribution of DocLayNet pages across document categories.  
}\label{fig:doc_categories_pie} 
\end{figure}

The pages in DocLayNet can be grouped into six distinct categories, namely \textit{Financial Reports}, \textit{Manuals}, \textit{Scientific Articles}, \textit{Laws \& Regulations}, \textit{Patents} and \textit{Government Tenders}. Each document category was sourced from various repositories. For example, Financial Reports contain both \textit{free-style} format annual reports\footnote{e.g. AAPL from https://www.annualreports.com/} which expose company-specific, artistic layouts as well as the more formal SEC filings. The two largest categories (\textit{Financial Reports} and \textit{Manuals}) contain a large amount of free-style layouts in order to obtain maximum variability. In the other four categories, we boosted the variability by mixing documents from independent providers, such as different government websites or publishers. In Figure~\ref{fig:doc_categories_pie}, we show the document categories contained in DocLayNet with their respective sizes.

We did not control the document selection with regard to language. The vast majority of documents contained in DocLayNet (close to  95\%) are published in English language. However, DocLayNet also contains a number of documents in other languages such as German (2.5\%), French (1.0\%) and Japanese (1.0\%). While the document language has negligible impact on the performance of computer vision methods such as object detection and segmentation models, it might prove challenging for layout analysis methods which exploit textual features.

To ensure that future benchmarks in the document-layout analysis community can be easily compared, we have split up DocLayNet into pre-defined train-, test- and validation-sets. 
In this way, we can avoid spurious variations in the evaluation scores due to random splitting in train-, test- and validation-sets. We also ensured that less frequent labels are represented in train and test sets in equal proportions. 

Table~\ref{tab:DLN} shows the overall frequency and distribution of the labels among the different sets. Importantly, we ensure that subsets are only split on full-document boundaries. This avoids that pages of the same document are spread over train, test and validation set, which can give an undesired evaluation advantage to models and lead to overestimation of their prediction accuracy. We will show the impact of this decision in Section~\ref{sec:results}.

In order to accommodate the different types of models currently in use by the community, we provide  DocLayNet in an \textit{augmented} COCO format~\cite{Lin-etal-2014-COCO}. This entails the standard COCO ground-truth file (in JSON format) with the associated page images (in PNG format, 1025$\times$1025 pixels). Furthermore, custom fields have been added to each COCO record to specify document category, original document filename and page number. In addition, we also provide the original PDF pages, as well as sidecar files containing parsed PDF text and text-cell coordinates (in JSON). All additional files are linked to the primary page images by their matching filenames.

Despite being cost-intense and far less scalable than automation, human annotation has several benefits over automated ground-truth generation.
The first and most obvious reason to leverage human annotations is the freedom to annotate any type of document without requiring a programmatic source. For most PDF documents, the original source document is not available. The latter is not a hard constraint with human annotation, but it is for automated methods. 
A second reason to use human annotations is that the latter usually provide a more natural interpretation of the page layout. The human-interpreted layout can significantly deviate from the programmatic layout used in typesetting. For example, ``invisible'' tables might be used solely for aligning text paragraphs on columns. Such typesetting tricks might be interpreted by automated methods incorrectly as an actual table, while the human annotation will interpret it correctly as  \textit{Text} or other styles. The same applies to multi-line text elements, when authors decided to space them as ``invisible'' list elements without bullet symbols. 
A third reason to gather ground-truth through human annotation is to estimate a ``natural'' upper bound on the segmentation accuracy. As we will show in Section~\ref{sec:DAC}, certain documents featuring complex layouts can have different but equally acceptable layout interpretations. This natural upper bound for segmentation accuracy can be found by annotating the same pages multiple times by different people and evaluating the inter-annotator agreement. Such a base-line consistency evaluation is very useful to define expectations for a good target accuracy in trained deep neural network models and avoid overfitting (see Table~\ref{tab:DLN}). On the flip side, achieving high annotation consistency proved to be a key challenge in human annotation, as we outline in Section~\ref{sec:DAC}.

\section{Annotation Campaign}\label{sec:DAC}
\vspace{3.5mm}

The annotation campaign was carried out in four phases. In phase one, we identified and prepared the data sources for annotation. In phase two, we determined the class labels and how annotations should be done on the documents in order to obtain maximum consistency. The latter was guided by a detailed requirement analysis and exhaustive experiments. In phase three, we trained the annotation staff and performed exams for quality assurance. In phase four, we distributed the annotation workload and performed continuous quality controls. Phase one and two required a small team of experts only. For phases three and four, a group of 40 dedicated annotators were assembled and supervised.

\begin{figure}[t!]
  \center
  \setlength{\fboxsep}{0pt}
  \fbox{\includegraphics[width=\linewidth]{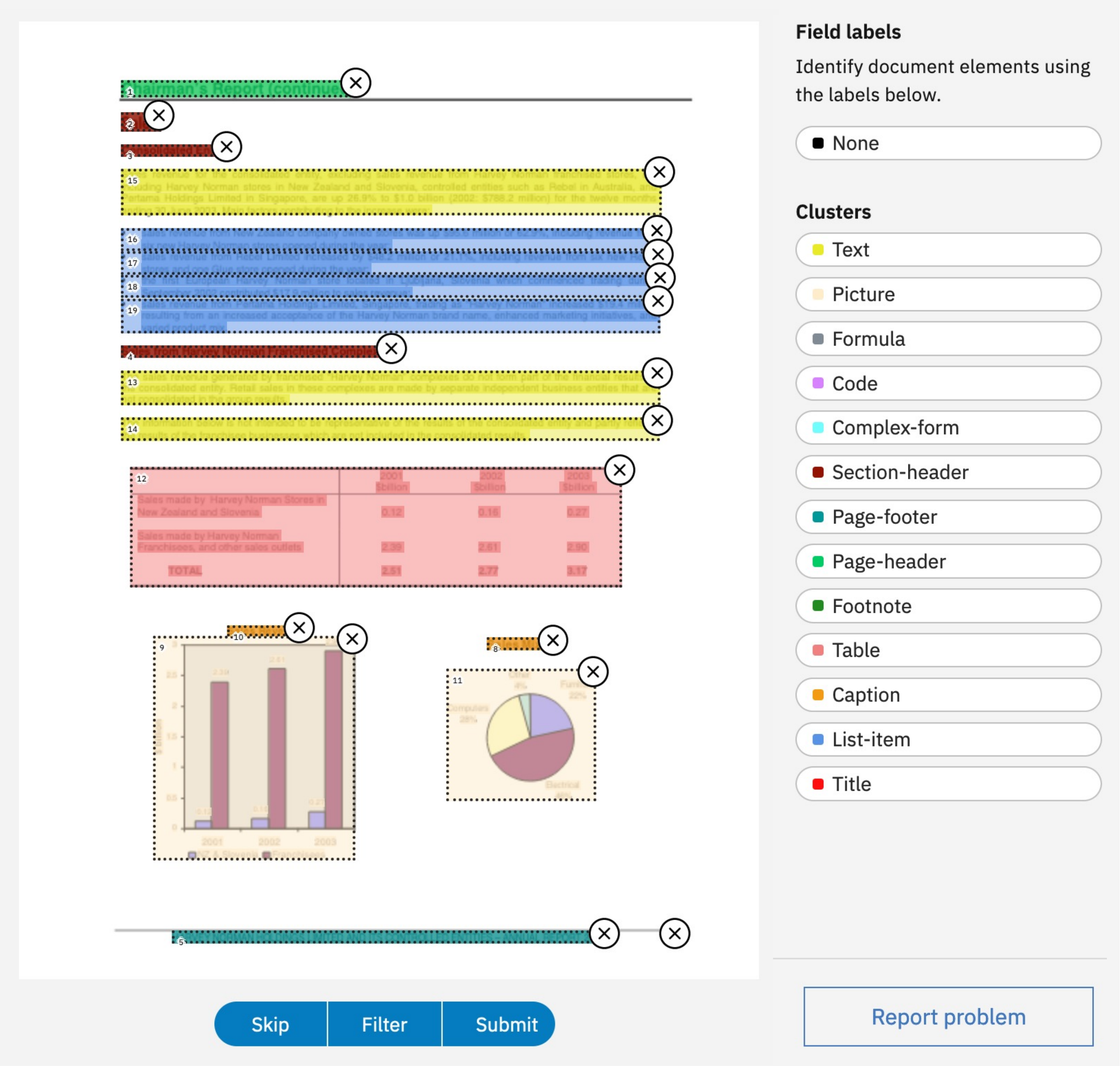}}
\caption{Corpus Conversion Service annotation user interface. The PDF page is shown in the background, with overlaid text-cells (in darker shades). The annotation boxes can be drawn by dragging a rectangle over each segment with the respective label from the palette on the right.
}\label{fig:annotator}
\end{figure}

\textbf{Phase 1: Data selection and preparation.} Our inclusion criteria for documents were described in Section~\ref{sec:DLN}. 
A large effort went into ensuring that all documents are free to use. The data sources include publication repositories such as arXiv\footnote{\url{https://arxiv.org/}}, government offices, company websites as well as data directory services for financial reports and patents. Scanned documents were excluded wherever possible because they can be rotated or skewed. This would not allow us to perform annotation with rectangular bounding-boxes and therefore complicate the annotation process.

Preparation work included uploading and parsing the sourced PDF documents in the Corpus Conversion Service (CCS)~\cite{Staar-etal-2018-CCS}, a cloud-native platform which provides a visual annotation interface and allows for dataset inspection and analysis. The annotation interface of CCS is shown in Figure~\ref{fig:annotator}. The desired balance of pages between the different document categories was achieved by selective subsampling of pages with certain desired properties. For example, we made sure to include the title page of each document and bias the remaining page selection to those with figures or tables. The latter was achieved by leveraging pre-trained object detection models from PubLayNet, which helped us estimate how many figures and tables a given page contains. 

\textbf{Phase 2: Label selection and guideline.} We reviewed the collected documents and identified the most common structural features they exhibit. This was achieved by identifying recurrent layout elements and lead us to the definition of 11 distinct class labels. These 11 class labels are \textit{Caption}, \textit{Footnote}, \textit{Formula}, \textit{List-item}, \textit{Page-footer}, \textit{Page-header}, \textit{Picture}, \textit{Section-header}, \textit{Table}, \textit{Text}, and \textit{Title}. Critical factors that were considered for the choice of these class labels were (1) the overall occurrence of the label, (2) the specificity of the label, (3) recognisability on a single page (i.e. no need for context from previous or next page) and (4) overall coverage of the page. Specificity ensures that the choice of label is not ambiguous, while coverage ensures that all meaningful items on a page can be annotated. We refrained from class labels that are very specific to a document category, such as \textit{Abstract} in the \textit{Scientific Articles} category. We also avoided class labels that are tightly linked to the semantics of the text. Labels such as \textit{Author} and \textit{Affiliation}, as seen in DocBank, are often only distinguishable by discriminating on the textual content of an element, which goes beyond visual layout recognition, in particular outside the \textit{Scientific Articles} category.

\begin{figure}[t!]
  \center
  \includegraphics[width=\linewidth]{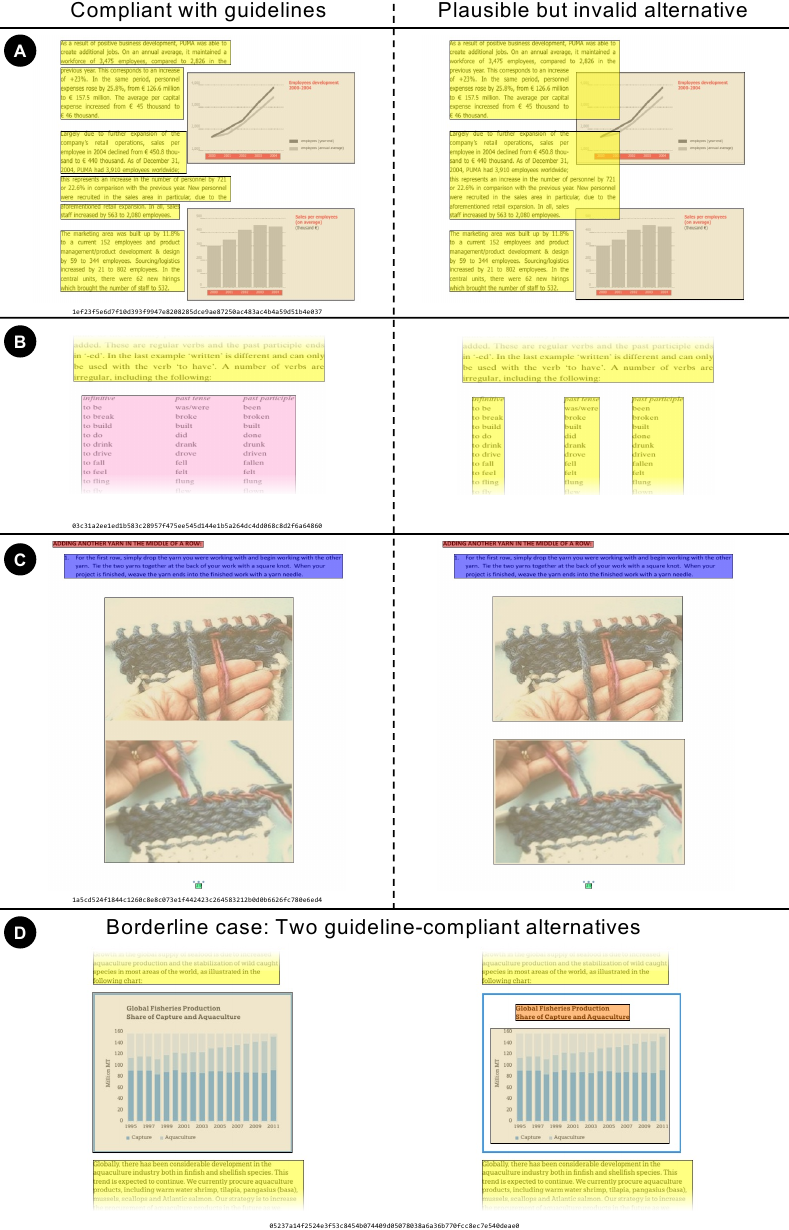}
\caption{Examples of plausible annotation alternatives for the same page. Criteria in our annotation guideline can resolve cases A to C, while the case D remains ambiguous.}
\label{fig:Plau_versus_Inc}
\end{figure}

At first sight, the task of visual document-layout interpretation appears intuitive enough to obtain plausible annotations in most cases. 
However, during early trial-runs in the core team, we observed many cases in which annotators use different annotation styles, especially for documents with challenging layouts. For example, if a figure is presented with subfigures, one annotator might draw a single figure bounding-box, while another might annotate each subfigure separately. The same applies for lists, where one might annotate all list items in one block or each list item separately. In essence, we observed that challenging layouts would be annotated in different but plausible ways. To illustrate this, we show in Figure~\ref{fig:Plau_versus_Inc} multiple examples of plausible but inconsistent annotations on the same pages.

Obviously, this inconsistency in annotations is not desirable for datasets which are intended to be used for model training. To minimise these inconsistencies, we created a detailed annotation guideline. While perfect consistency across 40 annotation staff members is clearly not possible to achieve, we saw a huge improvement in annotation consistency after the introduction of our annotation guideline. A few selected, non-trivial highlights of the guideline are:
\begin{enumerate}
\item Every list-item is an individual object instance with class label \textit{List-item}. This definition is different from PubLayNet and DocBank, where all list-items are grouped together into one \textit{List} object.
\item A \textit{List-item} is a paragraph with hanging indentation. Single-line elements can qualify as \textit{List-item} if the neighbour elements expose hanging indentation. Bullet or enumeration symbols are not a requirement.
\item For every \textit{Caption}, there must be exactly one corresponding \textit{Picture} or \textit{Table}.
\item Connected sub-pictures are grouped together in one \textit{Picture} object.
\item Formula numbers are included in a \textit{Formula} object.
\item Emphasised text (e.g. in italic or bold) at the beginning of a paragraph is not considered a \textit{Section-header}, unless it appears exclusively on its own line.
\end{enumerate}
The complete annotation guideline is over 100 pages long and a detailed description is obviously out of scope for this paper. Nevertheless, it will be made publicly available alongside with DocLayNet for future reference.

\textbf{Phase 3: Training.} After a first trial with a small group of people, we realised that providing the annotation guideline and a set of random practice pages did not yield the desired quality level for layout annotation. Therefore we prepared a subset of pages with two different complexity levels, each with a practice and an exam part. 974 pages were reference-annotated by one proficient core team member. Annotation staff were then given the task to annotate the same subsets (blinded from the reference). By comparing the annotations of each staff member with the reference annotations, we could quantify how closely their annotations matched the reference. Only after passing two exam levels with high annotation quality, staff were admitted into the production phase. Practice iterations were carried out over a timeframe of 12 weeks, after which 8 of the 40 initially allocated annotators did not pass the bar.

\textbf{Phase 4: Production annotation.} The previously selected 80K pages were annotated with the defined 11 class labels by 32 annotators. This production phase took around three months to complete. All annotations were created online through CCS, which visualises the programmatic PDF text-cells as an overlay on the page. The page annotation are obtained by drawing rectangular bounding-boxes, as shown in Figure~\ref{fig:annotator}. With regard to the annotation practices, we implemented a few constraints and capabilities on the tooling level. 
First, we only allow non-overlapping, vertically oriented, rectangular boxes. For the large majority of documents, this constraint was sufficient and it speeds up the annotation considerably in comparison with arbitrary segmentation shapes. 
Second, annotator staff were not able to see each other's annotations. This was enforced by design to avoid any bias in the annotation, which could skew the numbers of the inter-annotator agreement (see Table~\ref{tab:DLN}). We wanted to avoid this at any cost in order to have clear, unbiased baseline numbers for human document-layout annotation. 
Third, we introduced the feature of \textit{snapping} boxes around text segments to obtain a pixel-accurate annotation and again reduce time and effort. The CCS annotation tool automatically shrinks every user-drawn box to the minimum bounding-box around the enclosed text-cells for all purely text-based segments, which excludes only \textit{Table} and \textit{Picture}. For the latter, we instructed annotation staff to minimise inclusion of surrounding whitespace while including all graphical lines. A downside of snapping boxes to enclosed text cells is that some wrongly parsed PDF pages cannot be annotated correctly and need to be skipped. 
Fourth, we established a way to flag pages as \textit{rejected} for cases where no valid annotation according to the label guidelines could be achieved. Example cases for this would be PDF pages that render incorrectly or contain layouts that are impossible to capture with non-overlapping rectangles. Such rejected pages are not contained in the final dataset.
With all these measures in place, experienced annotation staff managed to annotate a single page in a typical timeframe of 20s to 60s, depending on its complexity.

\begin{table}[t!]
\center
\caption{Prediction performance (mAP@0.5-0.95) of object detection networks on DocLayNet test set. The MRCNN (Mask R-CNN) and FRCNN (Faster R-CNN) models with ResNet-50 or ResNet-101 backbone were trained based on the network architectures from the \textit{detectron2} model zoo (Mask R-CNN {R50, R101}-FPN 3x, Faster R-CNN R101-FPN 3x), with default configurations. The YOLO implementation utilized was YOLOv5x6~\protect\cite{Yolov5}. All models were initialised using pre-trained weights from the COCO 2017 dataset.}\label{tab:ML4DLN}
\begin{tabular}{|l|c|cccc|}
\hline
               & human & \multicolumn{2}{c}{MRCNN} & FRCNN & YOLO  \\
               &       &  R50    & R101            & R101  & v5x6  \\ \hline
Caption        & 84-89 & 68.4    & 71.5            & 70.1  & 77.7  \\
Footnote       & 83-91 & 70.9    & 71.8            & 73.7  & 77.2  \\
Formula        & 83-85 & 60.1    & 63.4            & 63.5  & 66.2  \\
List-item      & 87-88 & 81.2    & 80.8            & 81.0  & 86.2  \\
Page-footer    & 93-94 & 61.6    & 59.3            & 58.9  & 61.1  \\
Page-header    & 85-89 & 71.9    & 70.0            & 72.0  & 67.9  \\
Picture        & 69-71 & 71.7    & 72.7            & 72.0  & 77.1  \\
Section-header & 83-84 & 67.6    & 69.3            & 68.4  & 74.6  \\
Table          & 77-81 & 82.2    & 82.9            & 82.2  & 86.3  \\
Text           & 84-86 & 84.6    & 85.8            & 85.4  & 88.1  \\
Title          & 60-72 & 76.7    & 80.4            & 79.9  & 82.7  \\ \hline
All            & 82-83 & 72.4    & 73.5            & 73.4  & 76.8  \\ \hline
\end{tabular}
\end{table}

\section{Experiments}\label{sec:results}
\vspace*{3.5mm}

The primary goal of DocLayNet is to obtain high-quality ML models capable of accurate document-layout analysis on a wide variety of challenging layouts.
As discussed in Section~\ref{sec:RW}, object detection models are currently the easiest to use, due to the standardisation of ground-truth data in COCO format~\cite{Lin-etal-2014-COCO} and the availability of general frameworks such as \textit{detectron2}~\cite{detectron2}. Furthermore, baseline numbers in PubLayNet and DocBank were obtained using standard object detection models such as Mask R-CNN and Faster R-CNN. As such, we will relate to these object detection methods in this paper and leave the detailed evaluation of more recent methods mentioned in Section~\ref{sec:RW} for future work.


In this section, we will present several aspects related to the performance of object detection models on DocLayNet. Similarly as in PubLayNet, we will evaluate the quality of their predictions using mean average precision (mAP) with 10 overlaps that range from 0.5 to 0.95 in steps of 0.05 (mAP@0.5-0.95). These scores are computed by leveraging the evaluation code provided by the COCO API~\cite{Lin-etal-2014-COCO}. 


\begin{figure}[t!]
  \center
\includegraphics[width=\linewidth]{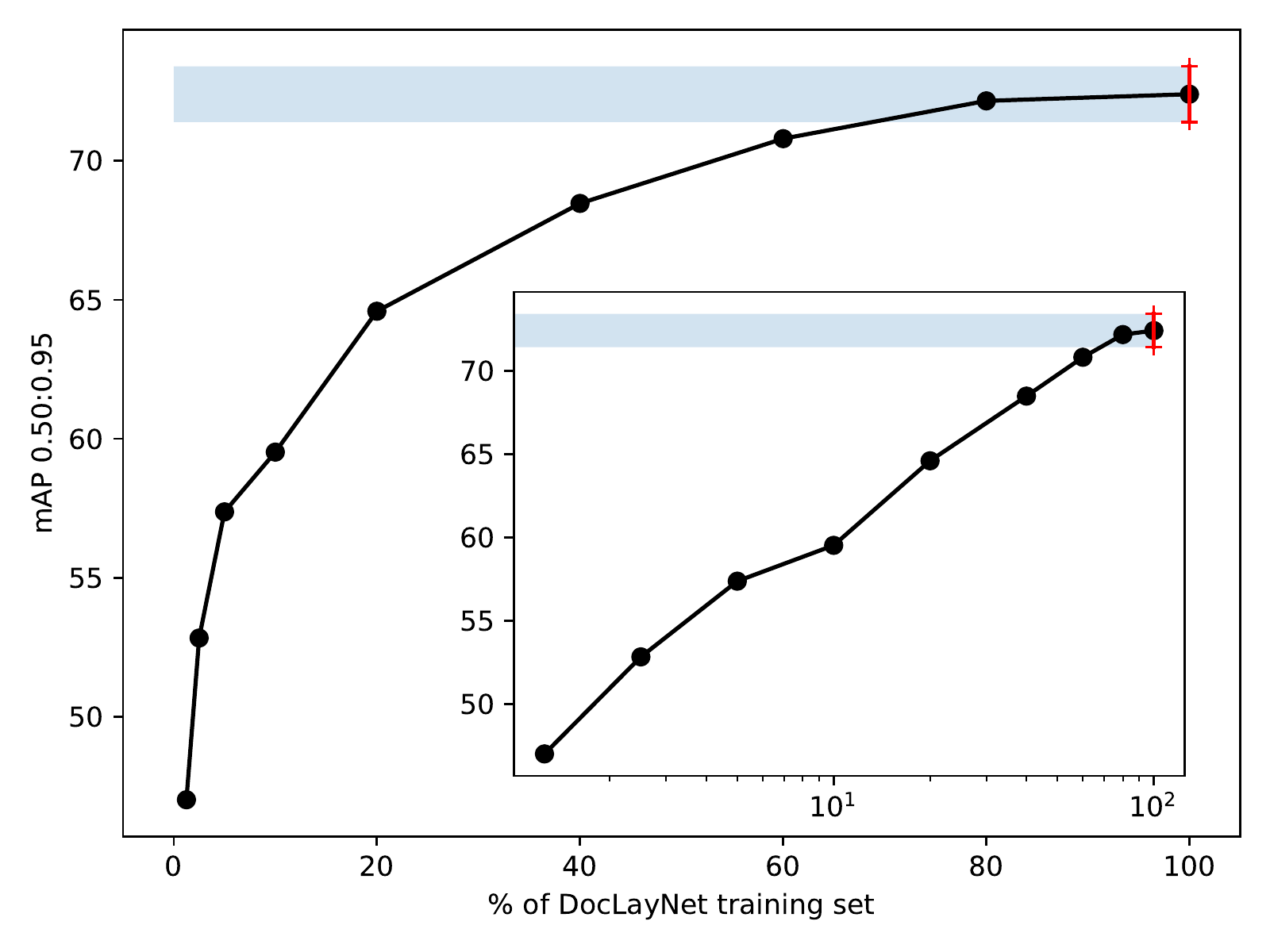}
\caption{Prediction performance (mAP@0.5-0.95) of a Mask R-CNN network with ResNet50 backbone trained on increasing fractions of the DocLayNet dataset. The learning curve flattens around the 80\% mark, indicating that increasing the size of the DocLayNet dataset with similar data will not yield significantly better predictions.}\label{fig:ablation}
\end{figure}

\subsection*{Baselines for Object Detection}
In Table~\ref{tab:ML4DLN}, we present baseline experiments (given in mAP) on Mask R-CNN~\cite{MaskRCNN}, Faster R-CNN~\cite{FasterRCNN}, and YOLOv5~\cite{Yolov5}. Both training and evaluation were performed on RGB images with dimensions of $1025\times1025$ pixels. 
For training, we only used one annotation in case of redundantly annotated pages. As one can observe, the variation in mAP between the models is rather low, but overall between 6 and 10\% lower than the mAP computed from the pairwise human annotations on triple-annotated pages. 
This gives a good indication that the DocLayNet dataset poses a worthwhile challenge for the research community to close the gap between human recognition and ML approaches. It is interesting to see that Mask R-CNN and Faster R-CNN produce very comparable mAP scores, indicating that pixel-based image segmentation derived from bounding-boxes does not help to obtain better predictions. On the other hand, the more recent Yolov5x model does very well and even out-performs humans on selected labels such as \textit{Text}, \textit{Table} and \textit{Picture}. This is not entirely surprising, as \textit{Text}, \textit{Table} and \textit{Picture} are abundant and the most visually distinctive in a document. 

\subsection*{Learning Curve} One of the fundamental questions related to any dataset is if it is ``large enough''. To answer this question for DocLayNet, we performed a data ablation study in which we evaluated a Mask R-CNN model trained on increasing fractions of the DocLayNet dataset.  As can be seen in Figure~\ref{fig:ablation}, the mAP score rises sharply in the beginning and eventually levels out. To estimate the error-bar on the metrics, we ran the training five times on the entire data-set. This resulted in a 1\% error-bar, depicted by the shaded area in Figure~\ref{fig:ablation}. In the inset of Figure~\ref{fig:ablation}, we show the exact same data-points, but with a logarithmic scale on the x-axis. As is expected, the mAP score increases linearly as a function of the data-size in the inset. The curve ultimately flattens out between the 80\% and 100\% mark, with the 80\% mark falling within the error-bars of the 100\% mark. This provides a good indication that the model would not improve significantly by yet increasing the data size. 
Rather, it would probably benefit more from improved data consistency (as discussed in Section~\ref{sec:DLN}), data augmentation methods~\cite{Shorten:2019uq}, or the addition of more document categories and styles.

\begin{table}[t!]
  \center
  \caption{Performance of a Mask R-CNN R50 network in mAP@0.5-0.95 scores trained on DocLayNet with different class label sets. The reduced label sets were obtained by either down-mapping or dropping labels. 
  }\label{tab:class_ablation}
  \vspace*{3.5mm}
  \begin{tabular}{|l|cccc|}
    \hline
Class-count                   &  11 &  6      & 5       & 4       \\ \hline
    Caption        &  68 & Text    & Text    & Text    \\
    Footnote       &  71 & Text    & Text    & Text    \\
    Formula        &  60 & Text    & Text    & Text    \\
    List-item      &  81 & Text    & 82      & Text    \\
    Page-footer    &  62 & 62      & -    & -    \\
    Page-header    &  72 & 68      & -    & -    \\
    Picture        &  72 & 72      & 72      & 72      \\
    Section-header &  68 & 67      & 69      & 68      \\
    Table          &  82 & 83      & 82      & 82      \\
    Text           &  85 & 84      & 84      & 84      \\
    Title          &  77 & Sec.-h. & Sec.-h. & Sec.-h. \\ \hline
    Overall            &  72 & 73      & 78      & 77      \\ \hline
  \end{tabular}
\end{table}


\subsection*{Impact of Class Labels} The choice and number of labels can have a significant effect on the overall model performance. Since PubLayNet, DocBank and DocLayNet all have different label sets, it is of particular interest to understand and quantify this influence of the label set on the model performance. We investigate this by either down-mapping labels into more common ones (e.g. \textit{Caption}$\rightarrow$\textit{Text}) or excluding them from the annotations entirely. Furthermore, it must be stressed that all mappings and exclusions were performed on the data before model training. In Table~\ref{tab:class_ablation}, we present the mAP scores for a Mask R-CNN R50 network on different label sets. Where a label is down-mapped, we show its corresponding label, otherwise it was excluded. We present three different label sets, with 6, 5 and 4 different labels respectively. The set of 5 labels contains the same labels as PubLayNet. However, due to the different definition of lists in PubLayNet (grouped list-items) versus DocLayNet (separate list-items), the label set of size 4 is the closest to PubLayNet, in the assumption that the \textit{List} is down-mapped to \textit{Text} in PubLayNet. The results in Table~\ref{tab:class_ablation} show that the prediction accuracy on the remaining class labels does not change significantly when other classes are merged into them. The overall macro-average improves by around 5\%, in particular when \textit{Page-footer} and \textit{Page-header} are excluded.

\begin{table}[t!]
  \center
  \caption{Performance of a Mask R-CNN R50 network with document-wise and page-wise split for different label sets. Naive page-wise split will result in \textapproxA{}10\% point improvement.}
  \label{tab:page_wise}
  \vspace*{3.5mm}
  \begin{tabular}{|l|cc|cc|}
    \hline
    Class-count & \multicolumn{2}{c|}{11} & \multicolumn{2}{c|}{5} \\
    Split &  Doc & Page & Doc & Page \\ \hline
    Caption        &  68     & 83   &     &      \\
    Footnote       &  71     & 84   &     &      \\
    Formula        &  60     & 66   &     &      \\
    List-item      &  81     & 88   & 82  & 88   \\
    Page-footer    &  62     & 89   &     &      \\
    Page-header    &  72     & 90   &     &      \\
    Picture        &  72     & 82   & 72  & 82   \\
    Section-header &  68     & 83   & 69  & 83   \\
    Table          &  82     & 89   & 82  & 90   \\
    Text           &  85     & 91   & 84  & 90   \\
    Title          &  77     & 81   &     &      \\ \hline
    All            &  72     & 84   & 78  & 87   \\ \hline
  \end{tabular}
\end{table}

\subsection*{Impact of Document Split in Train and Test Set} Many documents in DocLayNet have a unique styling. In order to avoid overfitting on a particular style, we have split the train-, test- and validation-sets of DocLayNet on document boundaries, i.e. every document contributes pages to only one set. To the best of our knowledge, this was not considered in PubLayNet or DocBank. To quantify how this affects model performance, we trained and evaluated a Mask R-CNN R50 model on a modified dataset version. Here, the train-, test- and validation-sets were obtained by a randomised draw over the individual pages. As can be seen in Table~\ref{tab:page_wise}, the difference in model performance is surprisingly large: page-wise splitting gains \~10\% in mAP over the document-wise splitting. Thus, random page-wise splitting of DocLayNet can easily lead to accidental overestimation of model performance and should be avoided.

\subsection*{Dataset Comparison} Throughout this paper, we claim that DocLayNet's wider variety of document layouts leads to more robust layout detection models. In Table~\ref{tab:DSvDS}, we provide evidence for that. We trained models on each of the available datasets (PubLayNet, DocBank and DocLayNet) and evaluated them on the test sets of the other datasets. Due to the different label sets and annotation styles, a direct comparison is not possible. Hence, we focussed on the common labels among the datasets. Between PubLayNet and DocLayNet, these are \textit{Picture}, \textit{Section-header}, \textit{Table} and \textit{Text}. Before training, we either mapped or excluded DocLayNet's other labels as specified in table~\ref{tab:class_ablation}, and also PubLayNet's \textit{List} to \textit{Text}. Note that the different clustering of lists (by list-element vs. whole list objects) naturally decreases the mAP score for \textit{Text}.

\begin{table}[t!]
\center
\caption{Prediction Performance (mAP@0.5-0.95) of a Mask R-CNN R50 network across the PubLayNet, DocBank \& DocLayNet data-sets. By evaluating on common label classes of each dataset, we observe that the DocLayNet-trained model has much less pronounced variations in performance across all datasets.}\label{tab:DSvDS}
\vspace*{3.5mm}
\begin{tabular}{|l|l|ccc|} \hline
                           &      & \multicolumn{3}{c|}{Testing on} \\
\multicolumn{1}{|c|}{Training on} & labels & PLN & DB & DLN \\ \hline
\multirow{5}{*}{PubLayNet (PLN)} & Figure  & 96 & 43  & 23  \\
                                 & Sec-header  & 87 & -   & 32  \\
                                 & Table   & 95 & 24  & 49  \\
                                 & Text    & 96 & -   & 42  \\
\cline{2-5}
                                 & total    & 93 & 34  & 30  \\ \hline
\multirow{3}{*}{DocBank (DB)}    & Figure  & 77 & 71  & 31  \\
                                 & Table   & 19 & 65  & 22  \\ \cline{2-5}
                                 & total   & 48 & 68  & 27  \\ \hline
\multirow{5}{*}{DocLayNet (DLN)} & Figure  & 67 & 51  & 72  \\
                                 & Sec-header  & 53 & -   & 68  \\
                                 & Table   & 87 & 43  & 82  \\
                                 & Text    & 77 & -   & 84  \\
\cline{2-5}
                                 & total   & 59 & 47  & 78  \\ \hline
\end{tabular}
\end{table}

\begin{figure*}[t!]
  \center
\includegraphics[width=\linewidth]{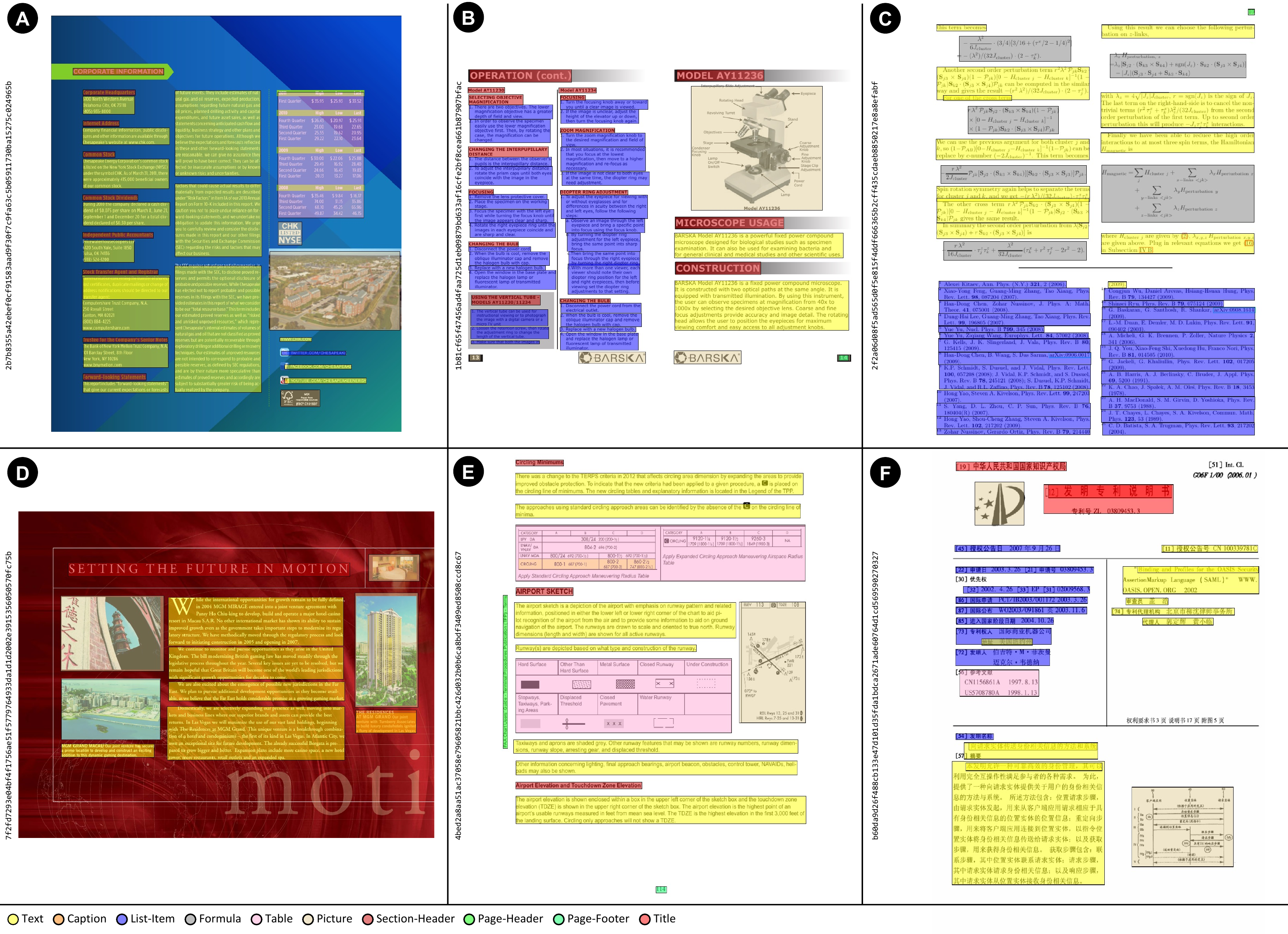}
\caption{Example layout predictions on selected pages from the DocLayNet test-set. (A, D) exhibit favourable results on coloured backgrounds. (B, C) show accurate list-item and paragraph differentiation despite densely-spaced lines. (E) demonstrates good table and figure distinction. (F) shows predictions on a Chinese patent with multiple overlaps, label confusion and missing boxes.
}\label{fig:segmentation_examples}
\end{figure*}

For comparison of DocBank with DocLayNet, we trained only on \textit{Picture} and \textit{Table} clusters of each dataset. We had to exclude \textit{Text} because successive paragraphs are often grouped together into a single object in DocBank. This paragraph grouping is incompatible with the individual paragraphs of DocLayNet. 
As can be seen in Table~\ref{tab:DSvDS}, DocLayNet trained models yield better performance compared to the previous datasets. It is noteworthy that the models trained on PubLayNet and DocBank perform very well on their own test set, but have a much lower performance on the foreign datasets. While this also applies to DocLayNet, the difference is far less pronounced. Thus we conclude that DocLayNet trained models are overall more robust and will produce better results for challenging, unseen layouts.

\subsection*{Example Predictions}
To conclude this section, we illustrate the quality of layout predictions one can expect from DocLayNet-trained models by providing a selection of examples without any further post-processing applied. Figure~\ref{fig:segmentation_examples} shows selected layout predictions on pages from the test-set of DocLayNet. Results look decent in general across document categories, however one can also observe mistakes such as overlapping clusters of different classes, or entirely missing boxes due to low confidence.

\section{\label{sec:conclusion}Conclusion}

In this paper, we presented the DocLayNet dataset. It provides the document conversion and layout analysis research community a new and challenging dataset to improve and fine-tune novel ML methods on. In contrast to many other datasets, DocLayNet was created by human annotation in order to obtain reliable layout ground-truth on a wide variety of publication- and typesetting-styles. Including a large proportion of documents outside the scientific publishing domain adds significant value in this respect.


From the dataset, we have derived on the one hand reference metrics for human performance on document-layout annotation (through double and triple annotations) and on the other hand evaluated the baseline performance of commonly used object detection methods. We also illustrated the impact of various dataset-related aspects on model performance through data-ablation experiments, both from a size and class-label perspective. Last but not least, we compared the accuracy of models trained on other public datasets and showed that DocLayNet trained models are more robust. 

To date, there is still a significant gap between human and ML accuracy on the layout interpretation task, and we hope that this work will inspire the research community to close that gap.

 \bibliographystyle{unsrt}
\bibliography{ccs_new}

\end{document}